\documentclass[10pt,twocolumn,letterpaper]{article}

\usepackage{iccv}              

\usepackage{subcaption}
\usepackage{adjustbox}
\usepackage{multirow}
\usepackage{pifont}
\usepackage{tabularray}
\usepackage{booktabs}
\usepackage{tabularx}
\usepackage{float}
\usepackage{stfloats}
\usepackage[accsupp]{axessibility} 

\newcolumntype{C}{>{\centering\arraybackslash}X}

\usepackage{colortbl}  
\usepackage{xcolor}    
\usepackage{pgfplots}  
\usepackage{float}

\definecolor{lightred}{rgb}{1,0.8,0.8}   
\definecolor{lightgreen}{rgb}{0.8,1,0.8} 
\newcommand{\rankcolor}[3]{%
    \pgfmathparse{100*(#3-#2)/(#1-#2)}%
    \xdef\shadeval{\pgfmathresult}%
    \cellcolor{lightgreen!\shadeval!lightred} #3
}

\definecolor{iccvblue}{rgb}{0.21,0.49,0.74}
\usepackage[pagebackref,breaklinks,colorlinks,allcolors=iccvblue]{hyperref}

\newcommand{\projectname}[0]{LayeredDepth}

\def\paperID{2616} 
\def\confName{ICCV}
\def\confYear{2025}

\title{Seeing and Seeing Through the Glass: \\ Real and Synthetic Data for Multi-Layer Depth Estimation}

\author{
Hongyu Wen \quad Yiming Zuo \quad  
Venkat Subramanian \quad  Patrick Chen \quad 
Jia Deng \\
Department of Computer Science, Princeton University \\
{\tt\small \{hongyu.wen,zuoym,venkat.subra,patrickchen,jiadeng\}@princeton.edu}}

\begin{document}
\maketitle

\begin{abstract}
Transparent objects are common in daily life, and understanding their multi-layer depth information—perceiving both the transparent surface and the objects behind it—is crucial for real-world applications that interact with transparent materials.
In this paper, we introduce LayeredDepth, the first dataset with multi-layer depth annotations, including a real-world benchmark and a synthetic data generator, to support the task of multi-layer depth estimation. Our real-world benchmark consists of 1,500 images from diverse scenes, and evaluating state-of-the-art depth estimation methods on it reveals that they struggle with transparent objects. The synthetic data generator is fully procedural and capable of providing training data for this task with an unlimited variety of objects and scene compositions. Using this generator, we create a synthetic dataset with 15,300 images. Baseline models training solely on this synthetic dataset produce good cross-domain multi-layer depth estimation. Fine-tuning state-of-the-art single-layer depth models on it substantially improves their performance on transparent objects, with quadruplet accuracy on our benchmark increased from 55.14\% to 75.20\%. All images and validation annotations are available under CC0 at \href{https://layereddepth.cs.princeton.edu}{https://layereddepth.cs.princeton.edu}.

\end{abstract}
\vspace{-1em}

\section{Introduction}
\label{sec:intro}

\begin{figure*}[t]
    \includegraphics[width=\linewidth]{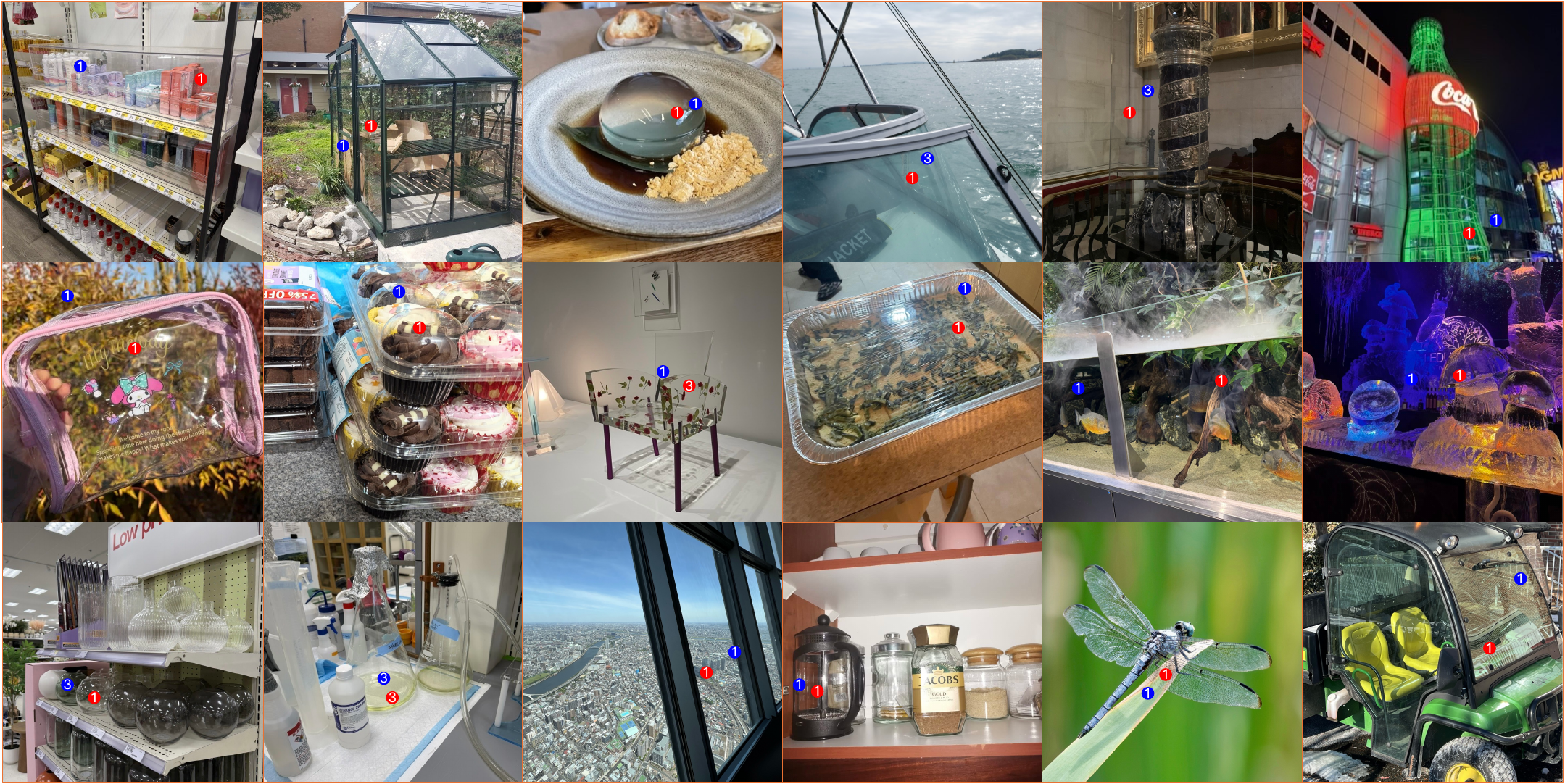  }
    \vspace{-1.5em}
    \caption{Gallery of our real-world benchmark \projectname{} along with sample relative depth pairs. Our benchmark comprises 1,500 images (available under CC0) from diverse scenes, including households, restaurants, laboratories, urban environments, and more. In the images, red and blue points indicate relative depth pairs, with red indicating a smaller depth and blue a larger one. The numbers on the points specify the annotated layer, for these examples, 1 means the frontmost surface and 3 means objects right behind transparent surfaces. }
    \vspace{-1em}
    \label{fig:benchmark_gallery}
\end{figure*}

Transparent objects are common in daily life, and understanding them is crucial for many real-world applications, such as autonomous navigation, 3D reconstruction, and dexterous manipulation. 

For many tasks, it is equally important to see both the transparent surfaces themselves as well as the objects behind them—in other words, to perceive depth across multiple layers. For instance, without the ability to see through glass, simple tasks like retrieving items from transparent containers or recognizing a scene behind a window would become difficult. In contrast, without the ability to perceive the transparent surface itself, we might struggle to grasp a plastic bag or accidentally walk into glass doors and walls.

To achieve human-level understanding of transparent objects, a perception system must be capable of capturing multi-layer depth information. To this end, we introduce a novel task \textit{multi-layer depth estimation}, which aims to predict the depth for all visible surfaces on and behind transparent objects by taking a single RGB image as input.

Existing datasets do not support this task. First, existing datasets only have single-layer depth annotations. While some datasets \cite{kitti1, kitti2} define depth on the objects behind the transparent surface and others \cite{cleargrasp, clearpose, phocal, booster} define depth on the transparent surfaces themselves, they offer only a partial representation of the scene and do not capture the full visual and geometric complexity of transparent surfaces. Second, existing datasets either contain only a small number of transparent objects~\cite{nyu_v1, nyu_v2, hypersim, kitti1, tartanair, structured3d, scannet, sun-rgbd}, or are restricted to a narrow set of indoor environments and typically tabletop objects~\cite{cleargrasp, clearpose, phocal, booster, transcg, seeingglass}. This limited scope makes it difficult to train or evaluate the generalizability of depth estimation methods for a perception system's real-world understanding of transparent objects.

In this paper, we introduce a real and a synthetic dataset tailored to the multi-layer depth estimation task. The real dataset is for benchmark purposes, containing in-the-wild images with high-quality, human-annotated relative depth ground-truth. Complementary to the real-world benchmark, our synthetic dataset allows us to train good-performing models for multi-layer depth estimation.

Our real-world benchmark consists of 1,500 images of transparent objects collected from diverse environments, including households, retail spaces, restaurants, laboratories, urban environments, and art installations, under various lighting conditions. 
Since ground-truth numerical depth cannot be accurately obtained for transparent objects, let alone multi-layer depth, we turn to relative depth annotations instead. Relative depth provides rich information about 3D structures and serves as an effective evaluation metric. Moreover, human annotators excel at determining relative depth, as they can reliably judge which of two points is closer to the camera and provide accurate annotations.
In total, we generate 14.2M tuples for relative depth annotations. A gallery of our benchmark along with sample annotations is shown in \cref{fig:benchmark_gallery}. All images and validation annotations are available under CC0. Our benchmark is highly challenging for state-of-the-art depth estimation methods \cite{depthanything, depthanythingv2, depthpro, geowizard, marigold, zoedepth, unidepth, moge, metric3dv2, midas}, even when evaluated on the simplified task of predicting only the first visible layer. For example, Metric3D V2 \cite{metric3dv2} achieves just 55.14\% quadruplet accuracy, while Depth Anything V2 \cite{depthanythingv2}, the most accurate among them, reaches only 70.43\%.

\begin{figure*}[t]
    \includegraphics[width=\linewidth]{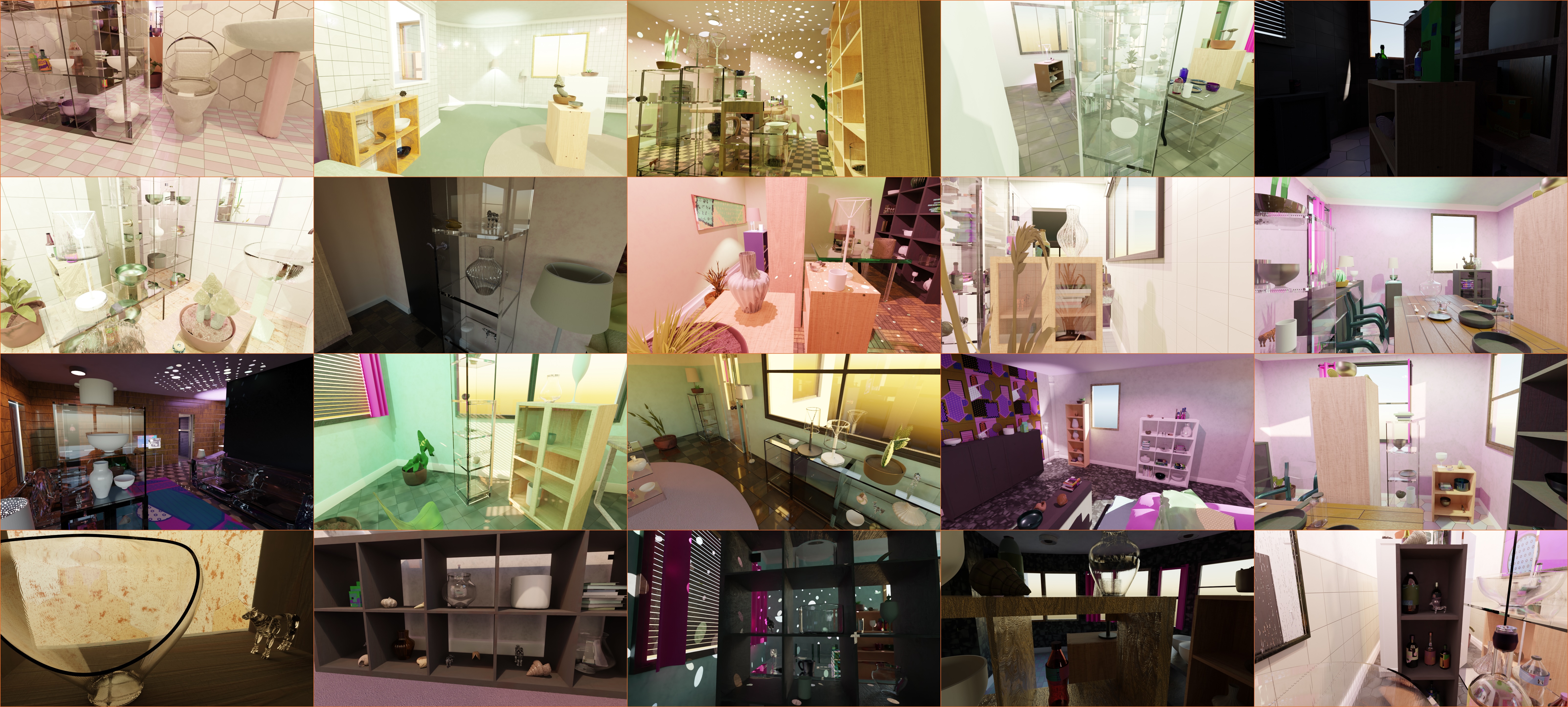}
    \caption{Gallery of our synthetic dataset. Our dataset is generated by \projectname{}-Syn, a procedural data generator that produces an unlimited diversity of shapes, materials, and spatial compositions.}
    \label{fig:syndata_gallery}
    \vspace{-0.5em}
\end{figure*}

\begin{figure*}[t]
    \centering
    \begin{subfigure}[t]{0.76\textwidth}
        \centering
        \adjustbox{valign=t}{\includegraphics[width=\linewidth]{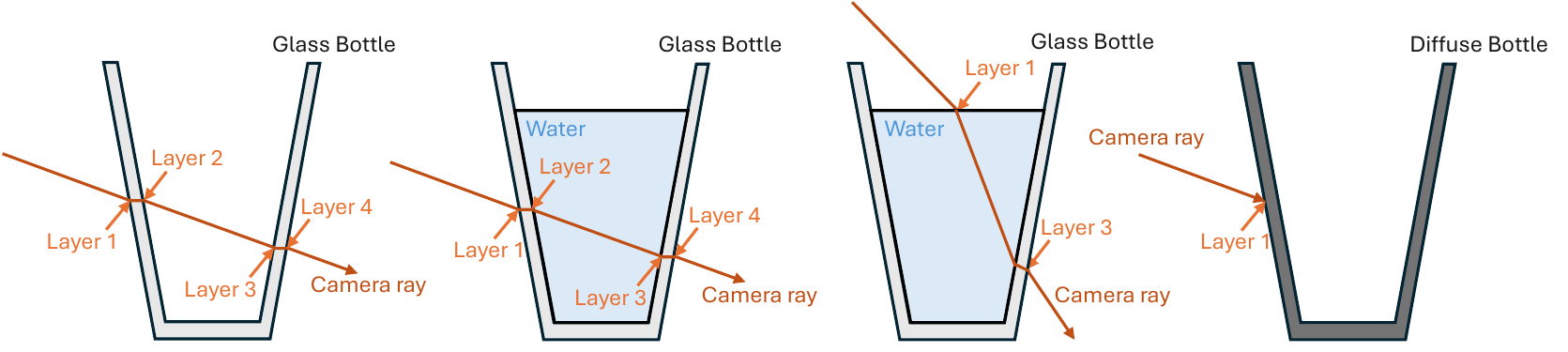}}
        \vspace{1.6em}
        \caption{}
        \label{fig:first}
    \end{subfigure}
    \hfill
    \begin{subfigure}[t]{0.22\textwidth}
        \centering
        \adjustbox{valign=t}{\includegraphics[width=\linewidth]{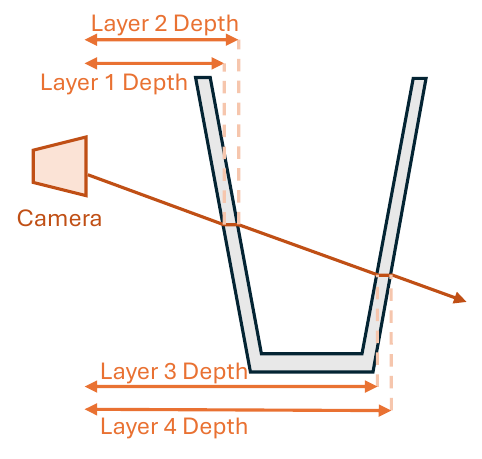}}
         \vspace{-0.5em}
        \caption{}
        \label{fig:second}
    \end{subfigure}
    \vspace{-0.9em}
    \caption{(a) Each transition in medium along the camera ray defines a distinct layer. (b) Depth on $i$-th layer is the distance along the z-axis from the $i$-th layer to the camera.}
    \label{fig:def_layer}
    \vspace{-0.9em}
\end{figure*}


For training, we introduce \projectname{}-Syn, a fully procedural synthetic data generator built on Infinigen Indoor \cite{infinigen_indoors}. It features a diverse library of procedural indoor assets with infinite variations in material, shape, and scene composition. To ensure the variety and frequency of transparent objects, our generator incorporates a random material assignment system, allowing any object to be designated as transparent.
Using this generator, we produce a synthetic dataset containing 15,300 images with multi-layer depth annotations. A gallery of our synthetic dataset is shown in \cref{fig:syndata_gallery}.
Training solely on this synthetic dataset, our baseline models design for multi-layer depth demonstrate strong cross-domain generalization, achieving promising results on real-world benchmarks. This highlights the quality of our dataset and marks an initial step toward addressing the multi-layer depth problem.
Moreover, fine-tuning state-of-the-art single-layer depth estimation model on our synthetic dataset leads to a substantial performance improvement on transparent objects, boosting quadruplet accuracy on our benchmark from 55.14\% to 75.20\%, further demonstrating the effectiveness of our synthetic data generator.

To summarize, our contributions are as follows:

\begin{itemize}
    \item We propose a new task, multi-layer depth estimation, and propose a baseline method to tackle this challenge.
    \item We propose a real-world multi-layer depth benchmark \projectname{} for transparent objects, consisting of 1,500 CC0 images of diverse scenes and 14.2M relative depth tuples. Our evaluation of state-of-the-art depth estimation methods on our benchmark reveals that they struggle significantly with transparent objects.
    \item We propose a procedural synthetic data generator \projectname{}-Syn for transparent objects and generate 15,300 images with multi-layer depth ground truth. Baseline models training solely on this synthetic dataset produce good cross-domain multi-layer depth estimation. Fine-tuning state-of-the-art depth models on it substantially improves their performance on transparent objects.
\end{itemize}

\section{Related Work}
\label{sec:related}

\paragraph{Real World Depth Datasets.}
Various real-world datasets have been proposed for depth prediction \cite{armeni2017joint, Matterport3d, scannet, scenenn, nyu_v1, nyu_v2, sun-rgbd, kitti1, kitti2, diode, cityscapes, megadepth, eth3d}.
These datasets typically employ structured light or time-of-flight (LiDAR) techniques. Because emitted light passes directly through transparent surfaces without sufficient reflections, these methods cannot generate reliable ground truth for transparent objects.

\paragraph{Synthetic Depth Datasets.}
Commonly used synthetic depth datasets \cite{sintel, hypersim, synthia, spring, irs, tartanair, structured3d, blendedmvs} do not specifically target transparent objects. They either lack transparent objects entirely or include only a few, some even lack accurate annotations. As a result, these datasets are insufficient for training models for transparent objects understanding.

\begin{figure}[t]
    \includegraphics[width=\linewidth]{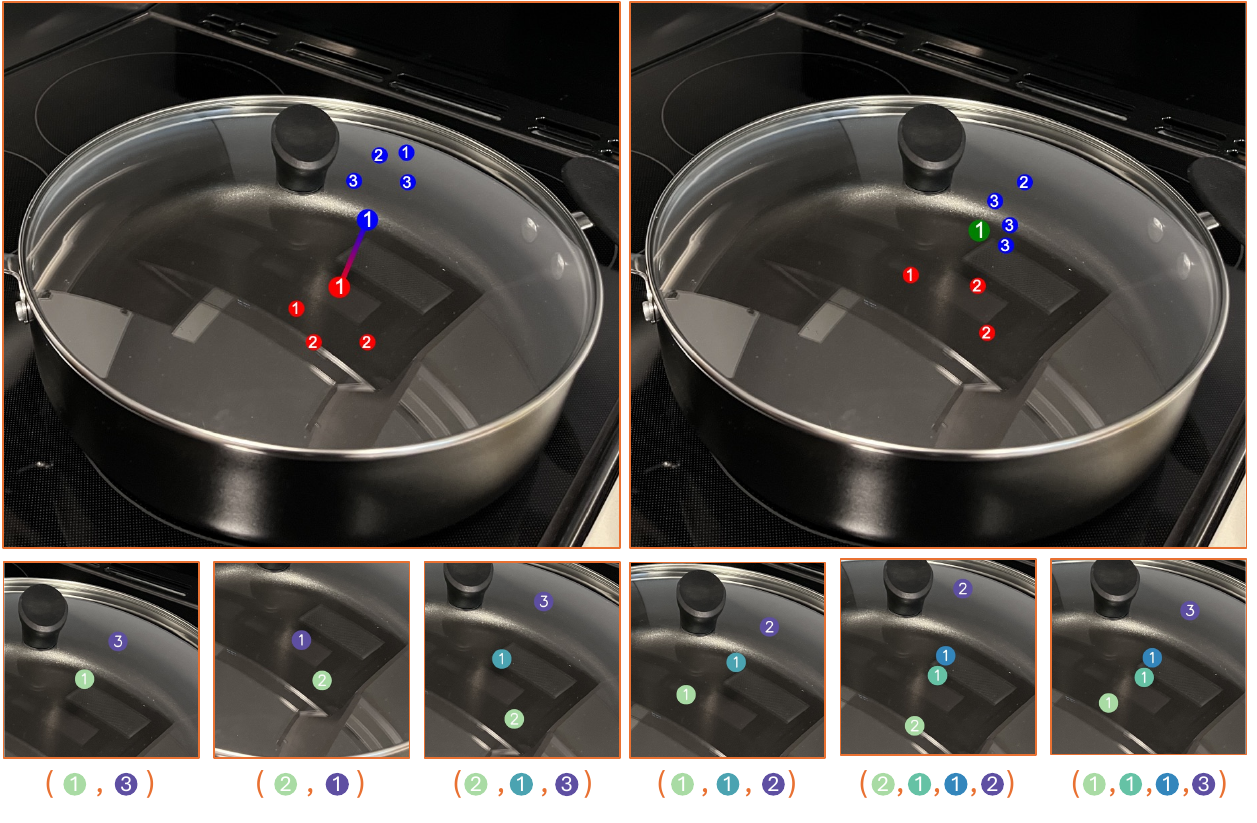}
    \caption{Our data annotation process for the real-world benchmark. The upper left image shows a monotonic depth line, along with points in front of the line (red) and points behind it (blue). The upper right image features a reference point (green) along with other points in front of and behind it. 
    The lower images display sampled relative depth tuples, where lefter elements in the tuple indicate smaller depths.
    The number assigned to each point corresponds to its respective depth layer. In this example, layer one represent the front side of a glass lid, layer two the back side of the lid, and layer three the interior of the pot.}
    \label{fig:benchmark_annotation}
    \vspace{-1em}
\end{figure}

\paragraph{Transparent Objects Benchmarks and Datasets} exist for various modalities \cite{transcut, trans10k, donthit, richcontext, tom-net, layeredflow}. Real-world depth benchmarks designed for transparent objects have been developed as well \cite{phocal, clearpose, transcg, seeingglass}. To obtain reliable depth ground-truth for transparent objects, they typically align 3D models of pre-scanned non-Lambertian objects with corresponding images. This approach restricts datasets to small objects that can be 3D scanned.
Booster \cite{booster} applies paint to non-Lambertian surfaces and employs structured lighting for stereo computation, which demands intensive manual labor and confines scenarios to indoor environments. Liang \etal \cite{glasswall} attach opaque patches onto glass walls and interpolate sparse measurements, limiting applicability to planar surfaces. \cite{seeing_through_the_glass, qiu2023looking} focus on predicting the 3D geometry behind glass, where the glass is typically a simple planar surface.
In contrast, our real-world benchmark covers a diverse range of scenes and objects.  

Similarly, existing synthetic datasets for transparent objects \cite{cleargrasp, lit, zhu2021rgb} are limited in scope, typically featuring desk-bound setups with restricted scene diversity. In contrast, our synthetic data generator is fully procedural and enabling unlimited object and scene compositions.

More importantly, none of existing depth benchmarks and datasets provided multi-layer annotations, which makes them inherently limited for transparent objects understanding. Our benchmark and dataset aim to support multi-layer depth task and provide multi-layer depth annotations.

\begin{figure*}[t]
    \includegraphics[width=\linewidth]{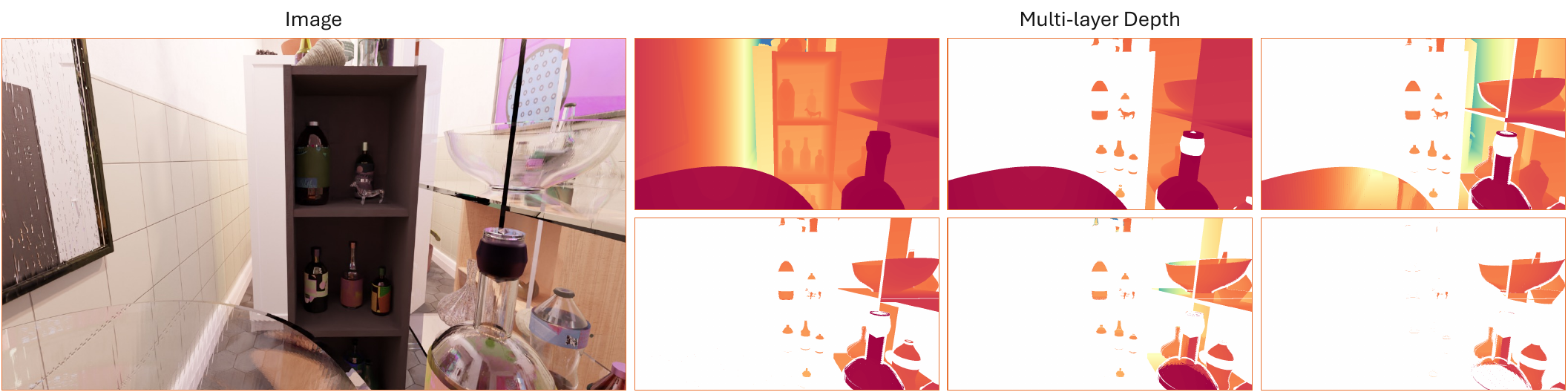}
    \caption{Showcase of our synthetic dataset with ground-truth annotations. Left: a sample image. Right: multi-layer depth ground truth, with layer 1 to 6 arranged from left to right, top to bottom.}
    \label{fig:multi-layer_gt_showcase}
\end{figure*}

\begin{figure*}[t]
    \includegraphics[width=\linewidth]{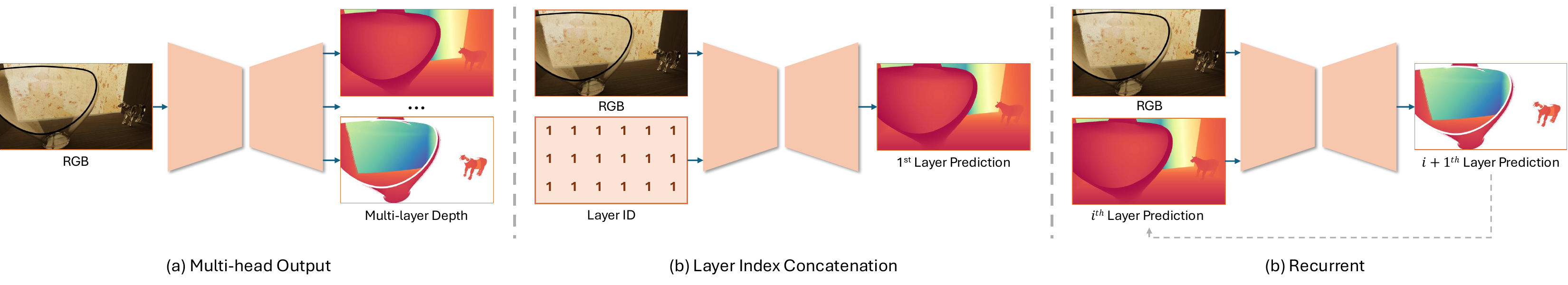}
    \caption{Our three multi-layer depth estimation baseline model design: Multi-head Output, Layer Index Concatenation and Recurrent.}
    \label{fig:multi_layer_model_design}
    \vspace{-0.5em}
\end{figure*}

\vspace{-0.3em}
\section{Multi-layer Depth}
\vspace{-0.3em}

When looking at transparent objects, humans naturally perceive the presence of multiple surfaces at various depths, including both points on the transparent surface itself and points on any occluded objects behind the transparent surface. To allow computer vision systems to develop this same understanding, we introduce the concept of \emph{layers}. In an image, each pixel corresponds to a camera ray, and every transition from one physical medium to another (e.g., from air to water) along the ray defines a distinct layer.

In the multi-layer depth prediction task, the goal is to predict the depth for each layer. More specifically, given an image $\mathcal{I}$ of resolution $H \times W$ and a query pixel $p = (x,y)$, the objective is to generate an ordered sequence of per-layer depth predictions $\hat{\mathcal{D}} = \{\hat{d}_1, \dots, \hat{d}_n\}$, where the number of layers $n$ may vary based on the query pixel. Here, $\hat{d}_i$ denotes the distance along the z-axis from the $i$-th layer to the camera. Some examples are shown in \cref{fig:def_layer}.


\vspace{-0.3em}
\section{\projectname{} Benchmark}
\vspace{-0.3em}
For our \projectname{} real-world benchmark, we aim to provide multi-layer depth ground-truth with a diverse coverage of different objects and scenes. We crowdsource images and manually filter them to ensure quality and diversity.
For ground-truth annotation, since numerical multi-layer depth cannot be accurately obtained for transparent objects, we turn to relative depth annotations instead.


\label{sec:benchmark}
\subsection{Image Acquisition}
We collected a dataset of 1,500 images featuring transparent objects, with 956 sourced through Prolific \cite{Prolific}. Each image was manually inspected to ensure the presence of distinct transparent surface features. The dataset captures a diverse range of scenes under various lighting conditions, including households, restaurants, laboratories, outdoor and urban environments, retail spaces, and car interiors. The transparent objects span four different materials, glass, plastic, liquid, and ice, including structural elements (e.g., glass walls, doors, staircases), household items (e.g., knives, pots, glass bottles, tables, plates, refrigerators), laboratory equipment (e.g., beakers, tubes), as well as food, artworks and buildings. A gallery of our benchmark is shown in \cref{fig:benchmark_gallery}.

\subsection{Annotation Acquisition}
Relative depth serves as an effective evaluation metric, and human annotators can provide these annotations reliably.

The annotators manually labeled all images using a custom interface, focusing on challenging areas with transparent surfaces, including clean, highly transparent materials, cluttered backgrounds, and strong reflections. 
They can annotate relative depth by drawing a monotonic depth line, along which depth increases consistently. Sampling points along the line will generate relative depth tuples such as pairs, triplets, or quadruplets. 
When no clear monotonic structure existed, the annotators used a simpler approach by selecting a reference point instead. In both ways, additional points could be placed in front of (smaller depth) or behind (larger depth) the whole depth line or the reference point, creating tuples of relative depth relationships across different surfaces. 
Each line and point was also labeled with a layer ID, to specify which layer was being annotated. An example of this process is shown in \cref{fig:benchmark_annotation}. To penalize models that generate excessive layers and false positives, we also include 10\% manually-labeled fake tuples representing nonexistent scenarios (e.g., a tuple labeled as layer 2 in a non-transparent region). During relative depth evaluation, these fake tuples will be marked incorrect if the model predicts any depth at those points.

Because multi-layer depth is not a familiar concept in daily life, achieving precise annotations through crowdsourcing is impractical. Therefore, we chose to annotate all images ourselves to ensure data quality. In total, we annotated 2.5M pairs, 5.9M triplets, and 5.8M quadruplets. 



\begin{table*}[t]
  \centering
  \begingroup
    \newcommand{\tub}[1]{\underline{\textbf{#1}}}
    \newcommand{\tb}[1]{\textbf{#1}}
    \newcommand{\tu}[1]{\underline{#1}}
  \resizebox{\linewidth}{!}{
\begin{tabular}{l c ccc c ccc c ccc c ccc c ccc c ccc}
    \toprule
        \multirow{2}{*}{Method} && 
        \multicolumn{3}{c}{All} && 
        \multicolumn{3}{c}{Mixed} && 
        \multicolumn{3}{c}{Layer 1} && 
        \multicolumn{3}{c}{Layer 3} && 
        \multicolumn{3}{c}{Layer 5} && \multicolumn{3}{c}{Layer 7} \\
        && P & T & Q && P & T & Q && P & T & Q 
        && P & T & Q && P & T & Q && P & T & Q \\
    \midrule
        Multi-head &
            & \tu{63.42} & \tu{42.55} & \tu{25.97}
            && \tu{74.72} & \tu{46.13} & \tu{26.38}
            && 65.66     & 39.93     & 24.21
            && \tb{56.58} & \tb{43.36} & \tb{30.77}
            && \tu{52.17} & \tu{38.66} & \tu{29.94}
            && \tu{38.40} & \tu{35.65} & \tu{33.10} \\

        Index Concat &
            & \tb{64.46} & \tb{44.00} & \tb{26.00}
            && \tb{76.70} & \tb{48.37} & \tb{26.46}
            && \tu{66.95} & \tu{41.84} & \tu{24.45}
            && \tu{55.85} & \tu{41.53} & \tu{30.06}
            && \tb{55.36} & \tb{42.69} & \tb{31.57}
            && \tb{39.66} & \tb{45.18} & \tb{45.08} \\

        Recurrent &
            & 62.36      & 41.88      & 24.64
            && 73.51      & 45.27      & 25.35
            && \tb{68.08} & \tb{44.46} & \tb{26.12}
            && 49.47      & 31.53      & 21.25
            && 45.51      & 30.06      & 21.29
            && 30.09      & 34.10      & 27.59 \\

    \bottomrule
\end{tabular}
  }
  \caption{Baseline methods evaluated on our real-world benchmark via tuple-wise accuracy. Best scores are in \textbf{bold}. Second best \underline{underlined}.}
  \label{tab:eval_real_baseline}
  \endgroup
\end{table*}

\begin{table*}[h]
  \centering
  \resizebox{\linewidth}{!}{
\begin{tabularx}{1.7\linewidth}{lCCCCCCCCCCCCCc}
    \toprule
            & ZoeDepth & UniDepth & GeoWizard & Marigold & MiDaS & MoGe & Metric3D & DA V1 & Depth4ToM & Tosi \emph{et~al.} & DA V2 & Depth Pro & Metric3D ft. \\
    \midrule
    P   & \rankcolor{89.53}{74.25}{74.25}
        & \rankcolor{89.53}{74.25}{77.03}
        & \rankcolor{89.53}{74.25}{81.39}
        & \rankcolor{89.53}{74.25}{82.59}
        & \rankcolor{89.53}{74.25}{76.61}
        & \rankcolor{89.53}{74.25}{76.76}
        & \rankcolor{89.53}{74.25}{80.31}
        & \rankcolor{89.53}{74.25}{78.02}
        & \rankcolor{89.53}{74.25}{84.44}
        & \rankcolor{89.53}{74.25}{84.72}
        & \rankcolor{89.53}{74.25}{85.34}
        & \rankcolor{89.53}{74.25}{87.39}
        & \rankcolor{89.53}{74.25}{89.53} \\
    T   & \rankcolor{81.71}{58.56}{58.56}
        & \rankcolor{81.71}{58.56}{62.15}
        & \rankcolor{81.71}{58.56}{66.29}
        & \rankcolor{81.71}{58.56}{68.35}
        & \rankcolor{81.71}{58.56}{62.05}
        & \rankcolor{81.71}{58.56}{63.99}
        & \rankcolor{81.71}{58.56}{65.43}
        & \rankcolor{81.71}{58.56}{62.95}
        & \rankcolor{81.71}{58.56}{74.50}
        & \rankcolor{81.71}{58.56}{73.80}
        & \rankcolor{81.71}{58.56}{74.44}
        & \rankcolor{81.71}{58.56}{76.29}
        & \rankcolor{81.71}{58.56}{81.71} \\
    Q   & \rankcolor{75.20}{52.43}{52.73}
        & \rankcolor{75.20}{52.43}{56.85}
        & \rankcolor{75.20}{52.43}{52.43}
        & \rankcolor{75.20}{52.43}{55.89}
        & \rankcolor{75.20}{52.43}{58.54}
        & \rankcolor{75.20}{52.43}{58.92}
        & \rankcolor{75.20}{52.43}{55.14}
        & \rankcolor{75.20}{52.43}{58.88}
        & \rankcolor{75.20}{52.43}{70.61}
        & \rankcolor{75.20}{52.43}{70.10}
        & \rankcolor{75.20}{52.43}{70.43}
        & \rankcolor{75.20}{52.43}{69.46}
        & \rankcolor{75.20}{52.43}{75.20} \\
    \bottomrule
\end{tabularx}
  }
  \caption{Depth methods evaluated on our real-world benchmark via tuple-wise accuracy, with greener color indicating better results.}
  \label{tab:realworld_firstlayer}
  \vspace{-1em}
\end{table*}

\section{\projectname{}-Syn Data Generator}

\subsection{Data Generator}
For model training, we seek help from synthetic data. Our synthetic data generator is built upon Infinigen Indoors \cite{infinigen_indoors}, a procedural system for generating photorealistic indoor scenes using Blender \cite{blender}. Infinigen Indoors synthesizes a wide variety of indoor objects, including furniture, appliances, cookware, dining utensils, architectural elements, and other common household items. Thanks to its procedural design, the generator can create endless variations at both the object and scene levels, resulting in unlimited diversity of shapes, materials, and spatial compositions.

To further curate the generator for transparent objects, we introduce several modifications to Infinigen Indoor:
\begin{itemize}
    \item We implement random material assignment system, allowing parts of objects to be altered to transparent materials such as glass. This significantly increases the frequency and diversity of transparent objects in our dataset.
    \item We relax certain scene arrangement constraints. While Infinigen Indoors is designed to produce aesthetically photorealistic environments, our training dataset benefits from more cluttered and varied spatial layouts. For instance, objects such as bowls and plates, which were previously restricted to kitchen settings, can now appear in any room. Similarly, Storage units and cabinets, formerly placed only against walls, may now be positioned freely, even in the middle of a room. This adjustment create more complex multi-layered scenes, where transparent objects may overlap, stack, or be embedded within intricate spatial configurations.
    \item We enhance lighting diversity by incorporating a new disco-style lighting system and adjusting outdoor lighting conditions. These modifications generate a broader range of illumination effects and introduce rich visual features on transparent surfaces. We also adjust the camera trajectory to include close-up, object-focused shots.
\end{itemize}

Using the generator, we create a synthetic dataset comprising 15,300 images, with 14,800 for training and 500 for validation. \cref{fig:syndata_gallery} showcases samples from our dataset.

\subsection{Multi-layer Ground Truth}
We provide multi-layer ground truth depth annotations alongside the images, as shown in \cref{fig:multi-layer_gt_showcase}. These annotations are aligned with the camera’s view, taking into account distortions caused by refraction rather than merely projecting object ground truth positions onto the imaging plane.

To obtain multi-layer depth ground-truth during rendering, we modified Blender’s ray tracing source code. Each ray is tracked as it moves through the scene. When it strikes a geometry surface and refracts, the corresponding layer is recorded, and its depth is logged.
Furthermore, to prevent reflected rays from transparent objects from contaminating the ground truth, we adjusted all transparent materials (such as glass and plastic) to be refraction-only and converted all other materials into diffuse surfaces during ground-truth rendering.


\section{Baseline Design}

We propose three baseline methods for the multi-layer depth prediction task as illustrated in \cref{fig:multi_layer_model_design}.
\begin{itemize}
    \item \textit{Multi-head Output} takes an RGB image as input and outputs multiple depth maps in a single forward pass.

    \item \textit{Layer Index Concatenation}: The layer ID concatenated with the RGB image forms a 4-channel input, prompting the model to output the depth of the corresponding layer.

    \item \textit{Recurrent}: The model iteratively predicts depth, taking as input the concatenation of the RGB image and the depth prediction of the previous layer. an all-zero tensor is used as the initial depth input.  
\end{itemize}
We adopt NeWCRFs~\cite{NeWCRFs} as the network backbone for its good metric depth performance and stable training.


\section{Experiments}
\label{sec:experiments}

\begin{figure*}[t]
    \includegraphics[width=\linewidth]{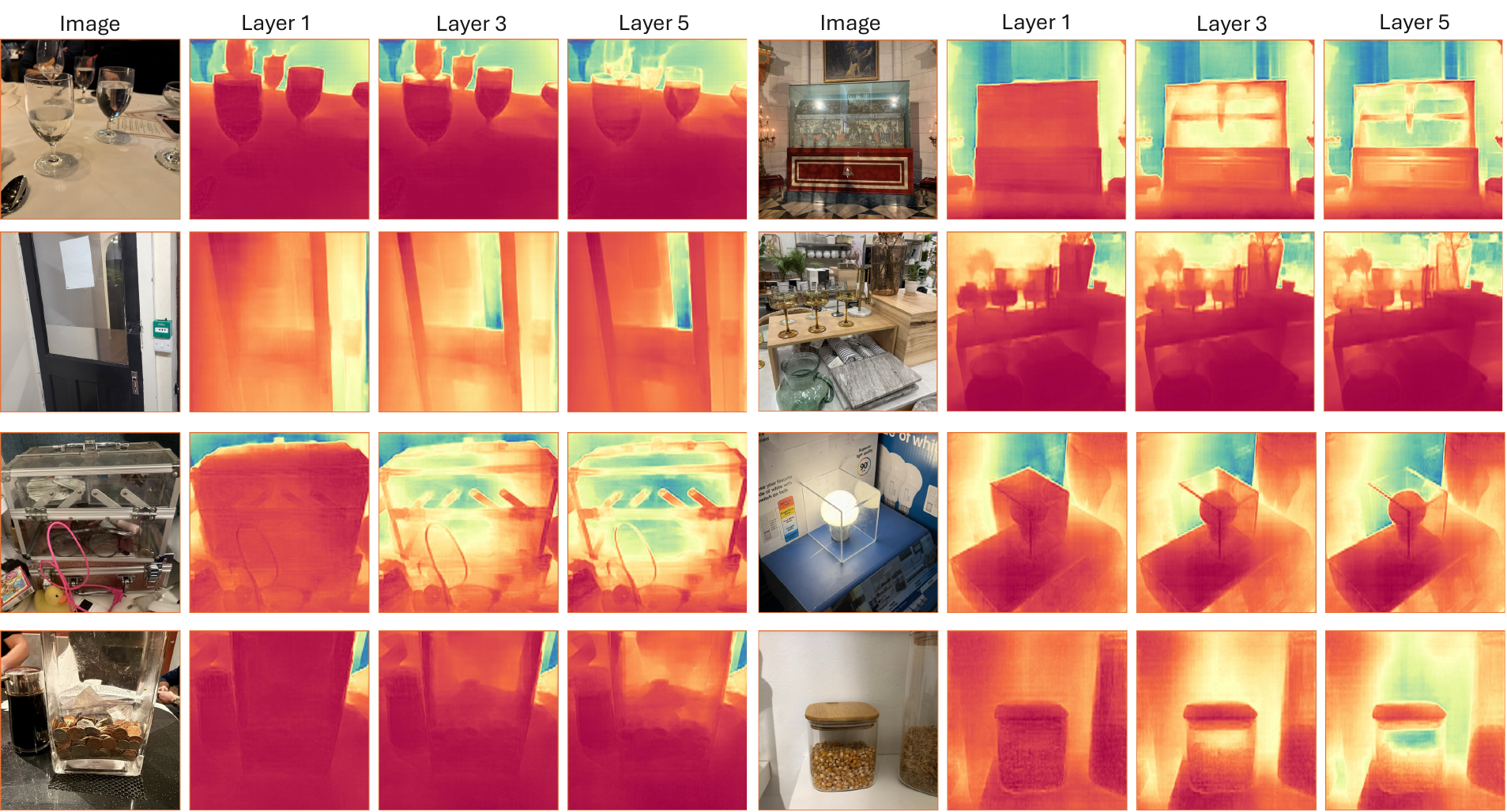}
    \caption{Visualizations of the multi-layer depth output of our Layer Index Concatenation baseline. Despite only being trained on our synthetic dataset, it shows impressive generalization to the real-world images in the wild. It generates consistent depth for opaque object, and can  progressively perceiving deeper layers as the layer ID increases, showing a strong spatial understanding of the transparent surfaces.}
    \label{fig:realworld_multilayer}
\end{figure*}

\subsection{Multi-Layer Depth Baseline Evaluation}
In this section, we evaluate our three baselines  for multi-layer depth on our real-world benchmark: Multi-head Output (Multi-head), Layer Index Concatenation (Index Concat), and Recurrent. We report tuple-wise accuracy for all tuple types: pairs (P), triplets (T), and quadruplets (Q), as well as six specific subsets: i) All: all the tuples, ii) Mixed: Tuples containing points from different layers. iii) Layer $i$: Tuples containing only points from layer $i$.
We report results only for odd-numbered layers ($i=1,3,5,7$), because in most cases, even-numbered layers have depths similar to the preceding odd-numbered layer. 

The results are shown in \cref{tab:eval_real_baseline}. Even when trained solely on our synthetic dataset, all three baseline models exhibit strong cross-domain generalization, achieving high accuracy. Among all the subsets, all three models achieve the highest accuracy ``Mixed'', as tuples in this category contain points from different surfaces, which often results in large depth differences, making them easier to distinguish.


Visualizations of Layer Index Concatenation baseline's results are shown in \cref{fig:realworld_multilayer}. Our baseline models demonstrate a strong spatial 3D understanding of transparent objects, progressively perceiving deeper layers as the layer ID increases.
However, there is still significant room for improvement. For example, the depth maps still exhibit some artifacts, particularly along object boundaries. But note that these baseline models are intended as proof-of-concept approaches and an initial step toward solving multi-layer depth estimation. We hope this work will inspire further research in this direction.

\subsection{Single-Layer Depth Experiments}
\begin{figure*}[t]
    \includegraphics[width=\linewidth]{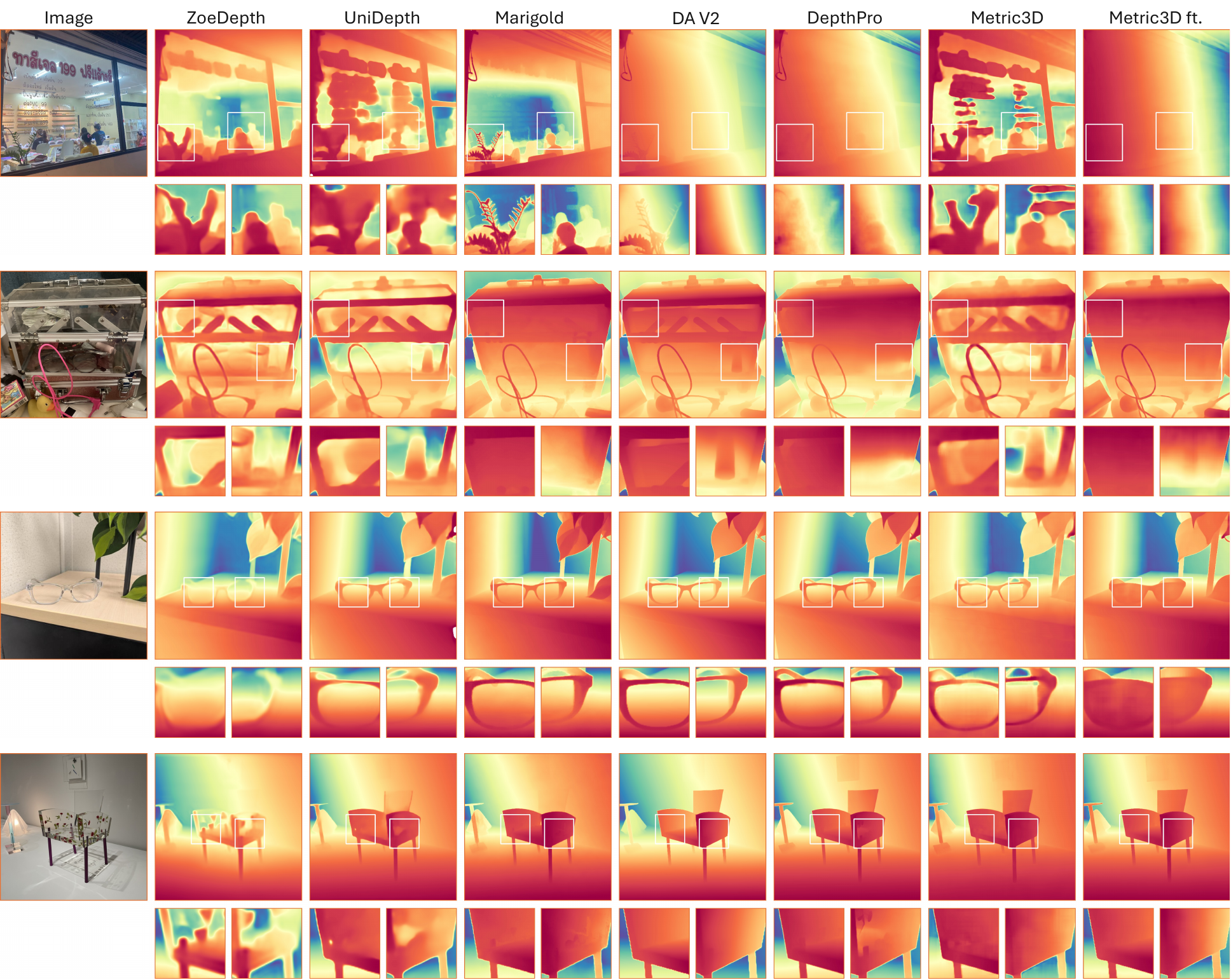}
    \caption{Qualitative comparison of state-of-the-art single layer depth methods on our benchmark. The original Metric3D produces blurry results, mixing the depth of multiple layers. While DepthPro and DA V2 generate finer results, they still face challenges in some cases (\eg, the glasses in the third row). Comparably, our fine-tuned Metric3D consistently generates high-quality depth maps.}
    \label{fig:realworld_firstlayer}
\end{figure*}

\begin{table*}[t]
  \centering
  \resizebox{\linewidth}{!}{
    \begin{tabularx}{1.5\textwidth}{l CCCC CCCC CCCC CCCC}
        \toprule
            \multirow{2}{*}{Method} 
            & \multicolumn{4}{c}{First Layer (all)} & \multicolumn{4}{c}{First Layer (trans)}
            & \multicolumn{4}{c}{Last Layer (trans)} & \multicolumn{4}{c}{Adapted Layer (trans)} \\
            & AbsRel$\downarrow$ & RMS$\downarrow$ & $\delta$1$\uparrow$ & $\delta$2$\uparrow$
            & AbsRel$\downarrow$ & RMS$\downarrow$ & $\delta$1$\uparrow$ & $\delta$2$\uparrow$ 
            & AbsRel$\downarrow$ & RMS$\downarrow$ & $\delta$1$\uparrow$ & $\delta$2$\uparrow$ 
            & AbsRel$\downarrow$ & RMS$\downarrow$ & $\delta$1$\uparrow$ & $\delta$2$\uparrow$ \\
        \midrule
        \multicolumn{17}{c}{Metric Depth} \\
        \midrule
        ZoeDepth      & 1.23   & 1.01   & 0.24      & 0.49    & 1.28   & 1.47   & 0.17    & 0.38    & 0.65   & 1.25   & 0.34    & 0.60    & 0.64   & 1.08   & 0.40    & 0.63 \\
        UniDepth V2   & 0.69   & 1.02   & 0.53      & 0.75    & 1.04   & 2.29   & 0.36    & 0.59    & 0.67   & 2.20   & 0.50    & 0.73    & 0.63   & 1.95   & 0.65    & 0.81 \\
        $\text{Metric3D V2}^{\S}$ 
                      & \underline{0.36}   & \underline{0.40}   & \textbf{0.69}   & \underline{0.85}  
                      & \underline{0.31}   & \textbf{0.50}   & \textbf{0.66}   & \underline{0.85}  
                      & \textbf{0.27}   & \textbf{0.96}   & \textbf{0.55}   & \underline{0.74}  
                      & \textbf{0.16}   & \textbf{0.37}   & \textbf{0.84}   & \textbf{0.94} \\
        DepthPro     
                      & \textbf{0.29}   & \textbf{0.36}   & \underline{0.69}   & \textbf{0.88}  
                      & \textbf{0.30}   & \underline{0.52}   & \underline{0.64}   & \textbf{0.86}  
                      & \underline{0.28}   & \underline{0.99}   & \underline{0.55}   & \textbf{0.76}  
                      & \underline{0.18}   & \underline{0.41}   & \underline{0.84}   & \underline{0.94} \\
        \midrule
        \multicolumn{17}{c}{Affine-invariant Depth} \\
        \midrule
        \midrule
        Marigold & 0.20 & 0.29 & 0.80 & 0.92 & 0.33 & 0.48 & 0.65 & 0.83 & 0.27 & 0.94 & 0.55 & 0.74 & 0.17 & 0.36 & 0.83 & 0.92 \\
        GeoWizard & 0.20 & 0.30 & 0.80 & 0.91 & 0.36 & 0.52 & 0.62 & 0.80 & \textbf{0.24} & \textbf{0.85} & \underline{0.61} & \textbf{0.80} & 0.16 & 0.34 & 0.85 & 0.93 \\
        $\text{MoGe}^{\dagger}$ & \underline{0.16} & 0.30 & \textbf{0.87} & \underline{0.94} & 0.43 & 0.66 & 0.69 & 0.84 & 0.35 & 1.05 & \textbf{0.64} & \underline{0.78} & 0.25 & 0.47 & \textbf{0.88} & \textbf{0.95} \\
        $\text{MiDaS}^{\ddagger}$ & 0.58 & 2.17 & 0.74 & 0.86 & 2.13 & 5.46 & 0.39 & 0.56 & 1.67 & 5.42 & 0.45 & 0.63 & 1.61 & 5.17 & 0.61 & 0.73 \\
        $\text{DA}^{\ddagger}$ & 0.73 & 2.23 & 0.79 & 0.89 & 2.32 & 5.33 & 0.39 & 0.59 & 1.80 & 5.30 & 0.49 & 0.66 & 1.75 & 5.06 & 0.65 & 0.75 \\
        $\text{DA V2}^{\ddagger}$ & 0.40 & 1.51 & 0.83 & 0.92 & 1.24 & 3.19 & 0.55 & 0.74 & 1.06 & 3.42 & 0.45 & 0.66 & 0.96 & 3.01 & 0.73 & 0.83 \\
        $\text{Depth4ToM}^{\ddagger}$ & 0.41 & 1.08 & 0.67 & 0.86 & 0.70 & 1.42 & 0.45 & 0.69 & 0.53 & 1.77 & 0.41 & 0.64 & 0.44 & 1.29 & 0.66 & 0.81 \\
        Tosi \emph{et~al.}           & 0.34 & 1.03 & 0.77 & 0.90 & 0.65 & 1.50 & 0.54 & 0.74 & 0.48 & 1.79 & 0.50 & 0.70 & 0.39 & 1.32 & 0.74 & 0.86 \\

        $\text{ZoeDepth}^{\P}$ & 0.28 & 0.43 & 0.74 & 0.86 & 0.58 & 0.84 & 0.46 & 0.67 & 0.32 & 0.94 & 0.57 & 0.77 & 0.26 & 0.59 & 0.73 & 0.85 \\
        $\text{UniDepth}^{\P}$ & 0.21 & 0.67 & 0.85 & 0.92 & 0.68 & 1.81 & 0.58 & 0.74 & 0.53 & 1.99 & 0.59 & 0.74 & 0.45 & 1.60 & 0.80 & 0.87 \\
        $\text{Metric3D V2}^{\P}$ & 0.16 & \underline{0.23} & 0.84 & 0.93 & \textbf{0.25} & \textbf{0.42} & \underline{0.70} & \underline{0.86} & \underline{0.25} & \underline{0.93} & 0.57 & 0.74 & \textbf{0.14} & \textbf{0.29} & 0.86 & 0.94 \\
        $\text{DepthPro}^{\P}$    & \textbf{0.14} & \textbf{0.22} & \underline{0.87} & \textbf{0.95} & \underline{0.25} & \underline{0.42} & \textbf{0.73} & \textbf{0.89} & 0.26 & 0.95 & 0.58 & 0.76 & \underline{0.14} & \underline{0.33} & \underline{0.88} & \underline{0.95} \\
        \bottomrule
    \end{tabularx}
  }
  \caption{
    Representative depth methods evaluated on synthetic validation set. Best scores are in \textbf{bold}. Second best \underline{underlined}.
    $\S$: Metric3D V2 predictions are scaled using ground-truth camera intrinsics.
    $\dagger$: MoGe is inherently a scale-invariant method, but we estimate an additional global shift for easier comparison with other affine-invariant methods.
    $\ddagger$: The predictions from MiDas, Depth Anything, Depth Anything V2, and Depth4ToM are aligned in disparity space.
    $\P$: ZoeDepth, UniDepth, Metric3D V2, and Depth Pro are metric depth methods but are evaluated in affine-invariant setting for a fair comparison.
  }
  \label{tab:syn_firstlayer}
  \vspace{-1.2em}
\end{table*}

We evaluate ten state-of-the-art depth estimation methods on our real-world benchmark, including Depth Anything (DA) \cite{depthanything}, Depth Anything V2 (DA V2) \cite{depthanythingv2}, Depth Pro \cite{depthpro}, ZoeDepth \cite{zoedepth}, Unidepth V2 \cite{unidepthv2}, GeoWizard \cite{geowizard}, Marigold \cite{marigold}, MiDaS V3.1 \cite{midas}, MoGe \cite{moge}, and Metric3D V2 \cite{metric3dv2}, and two methods tailored for transparent objects, including Tosi \emph{et~al.} \cite{tosi2024diffusion} and Depth4ToM \cite{depth4tom}.
To assess the effectiveness of our synthetic dataset, we also evaluate a depth model fine-tuned on our synthetic data. We choose Metric3D V2 \cite{metric3dv2} for fine-tuning, as it provides publicly available fine-tuning code (see supplementary for details).
Since existing depth models only perform single-layer depth estimation, we evaluate them exclusively on the Layer 1 subset, where models are only required to predict depth for the frontmost layer. Similarly, the fine-tuning is conducted solely on the Layer 1 of our synthetic dataset.

Qualitative and quantitative results are shown in \cref{fig:realworld_firstlayer} and \cref{tab:realworld_firstlayer}, respectively. Despite their strong zero-shot generalization on normal scenes, all state-of-the-art methods struggle when handling transparency.
The best-performing models, DA V2 and Depth Pro, achieve only 85.34\% and 87.39\% pair-wise accuracy, and 70.43\% and 69.46\% quadruplet-wise accuracy. Visualizations reveal that most methods fail on clean transparent surfaces, often producing blurry, artifact-ridden depth estimates that mix information from different layers. While Depth Pro and DA V2 are the most reliable, generating mostly smooth predictions, they still exhibit notable artifacts in some cases (e.g., DA V2 in the second row, Depth Pro in the fourth row) or completely fail to detect clean transparent surfaces (e.g., third row).

Even though Metric3D V2 is not the best-performing method among existing models, fine-tuning it on our synthetic dataset significantly improves its performance on transparent objects. Quadruplet accuracy increases from 55.14\% to 75.20\%, surpassing all previously reported results.
Visualizations further demonstrate that the fine-tuned Metric3D V2 consistently produces high-quality depth maps, even in cases where DA V2 and Depth Pro struggle. This clearly highlights the effectiveness of our synthetic data generator for transparent objects understanding.

To further provide numerical evaluation,  we conduct zero-shot evaluation of the state-of-the-art methods on our synthetic validation set. 
We use the relative point error (AbsRel), root mean square error (RMS) and the percentage of inliners $\delta_i$, $i \in \{ 1, 2\} $ with threshold $1.25^i$ as metrics.
We report performance across all pixels (all) and specifically on pixels corresponding to transparent objects (trans).
For transparent objects, as we do not know which layer a single-layer method is actually predicting, we compare their predictions against multi-layer ground truth using three strategies:
a) First Layer: Following real-world benchmark evaluations, predictions are compared against the ground truth of Layer 1.
b) Last Layer: Predictions are compared to the last visible surfaces, requiring the model to see through all transparent objects.
c) Adapted Layer: This approach allows the model to ``cheat'' by matching each predicted depth value to the closest depth layer in the ground truth.

Results are shown in \cref{tab:syn_firstlayer}. For metric depth, Metric3D and DepthPro achieve the best results, while for affine-invariant depth, DepthPro performs the best. 
Overall, all methods show significantly higher errors on our dataset compared to widely used depth benchmarks, particularly on transparent regions, highlighting the challenges of handling transparency in depth estimation.
Notably, the Adapted Layer strategy exhibits errors on transparent objects than the First or Last Layer strategies.
This aligns with our observation that existing methods often struggle to disentangle depth information from multiple layers. Rather than accurately predicting each layer, they tend to produce depth estimates that fluctuate between different depth layers at transparent regions. This observation further highlights the importance of multi-layer depth estimation, as the task inherently encourages models to disentangle conflicting multi-layer visual features more effectively.

\section{Conclusion}
We introduce \projectname{}, a real-world depth benchmark and a procedural synthetic data generator designed for multi-layer depth estimation of transparent objects. We believe our dataset will drive progress in this field.
\small

\newpage
\section*{Acknowledgments} This work was partially supported by the National Science Foundation. We additionally thank friends and colleagues at Princeton University for their data contribution.

\section{Appendix}

\begin{table*}[b!]
  \centering
  \begingroup
    \newcommand{\tub}[1]{\underline{\textbf{#1}}}
    \newcommand{\tb}[1]{\textbf{#1}}
    \newcommand{\tu}[1]{\underline{#1}}
  \resizebox{\linewidth}{!}{
\begin{tabular}{l c ccc c ccc c ccc c ccc c ccc c ccc}
    \toprule
        \multirow{2}{*}{Method} && 
        \multicolumn{3}{c}{All} && 
        \multicolumn{3}{c}{Mixed} && 
        \multicolumn{3}{c}{Layer 1} && 
        \multicolumn{3}{c}{Layer 3} && 
        \multicolumn{3}{c}{Layer 5} && \multicolumn{3}{c}{Layer 7} \\
        && P & T & Q && P & T & Q && P & T & Q 
        && P & T & Q && P & T & Q && P & T & Q \\
    \midrule
    Multi-head (NeWCRFs) &
        & 63.42       & 42.55       & 25.97
        && 74.72      & 46.13      & 26.38
        && 65.66      & 39.93      & 24.21
        && \tu{56.58} & \tu{43.36} & \tu{30.77}
        && 52.17      & 38.66      & 29.94
        && 38.40      & 35.65      & 33.10 \\

    Index Concat (NeWCRFs) &
        & \tu{64.46} & \tu{44.00} & \tu{26.00}
        && \tu{76.70} & \tu{48.37} & \tu{26.46}
        && 66.95      & 41.84      & 24.45
        && 55.85      & 41.53      & 30.06
        && \tu{55.36} & \tu{42.69} & \tu{31.57}
        && \tu{39.66} & \tu{45.18} & \tu{45.08} \\

    Recurrent (NeWCRFs) &
        & 62.36      & 41.88      & 24.64
        && 73.51      & 45.27      & 25.35
        && \tu{68.08} & \tu{44.46} & \tu{26.12}
        && 49.47      & 31.53      & 21.25
        && 45.51      & 30.06      & 21.29
        && 30.09      & 34.10      & 27.59 \\

    Multi-head (DA v2) &
        & \tb{80.72} & \tb{70.67} & \tb{61.68}
        && \tb{89.37} & \tb{72.98} & \tb{62.23}
        && \tb{85.80} & \tb{74.34} & \tb{65.62}
        && \tb{73.75} & \tb{62.68} & \tb{59.19}
        && \tb{66.19} & \tb{59.29} & \tb{54.75}
        && \tb{57.39} & \tb{50.78} & \tb{48.09} \\

    \bottomrule

\end{tabular}
  }
  \caption{Baseline methods evaluated on our real-world benchmark via tuple-wise accuracy. Best scores are in \textbf{bold}. Second best \underline{underlined}.}
  \label{tab:supp_eval_real_baseline}
  \endgroup
\end{table*}

\begin{table*}[b!]
  \centering
  \resizebox{\linewidth}{!}{
    \begin{tabularx}{1.5\textwidth}{l CCCC CCCC CCCC CCCC}
        \toprule
            \multirow{2}{*}{Method} 
            & \multicolumn{4}{c}{Layer 1} & \multicolumn{4}{c}{Layer 3}
            & \multicolumn{4}{c}{Layer 5} & \multicolumn{4}{c}{Layer 7} \\
            & AbsRel$\downarrow$ & RMS$\downarrow$ & $\delta$1$\uparrow$ & $\delta$2$\uparrow$
            & AbsRel$\downarrow$ & RMS$\downarrow$ & $\delta$1$\uparrow$ & $\delta$2$\uparrow$ 
            & AbsRel$\downarrow$ & RMS$\downarrow$ & $\delta$1$\uparrow$ & $\delta$2$\uparrow$ 
            & AbsRel$\downarrow$ & RMS$\downarrow$ & $\delta$1$\uparrow$ & $\delta$2$\uparrow$ \\
    \toprule
Multi-head (NeWCRFs)
  & 17.97 & \underline{23.41} & \underline{83.08} & 93.14
  & \underline{15.90} & \underline{47.39} & \underline{80.15} & \underline{93.89}
  & \underline{14.58} & \underline{47.51} & \underline{81.48} & \underline{94.12}
  & \underline{16.27} & \underline{58.78} & \underline{80.44} & \underline{93.92} \\

Index Concat (NeWCRFs)
  & 17.26 & 23.70 & 83.02 & 93.27
  & 16.25 & 48.19 & 79.63 & 93.72
  & 15.03 & 48.11 & 80.61 & 93.97
  & 16.49 & 59.03 & 80.26 & 93.76 \\

Recurrent (NeWCRFs)
  & \underline{17.23} & 25.53 & 81.89 & \underline{93.37}
  & 17.61 & 51.79 & 76.39 & 92.53
  & 17.25 & 52.53 & 76.54 & 92.20
  & 18.35 & 61.98 & 77.04 & 92.40 \\

Multi-head (DA v2)
  & \textbf{8.24}  & \textbf{15.21} & \textbf{93.43} & \textbf{97.73}
  & \textbf{10.68} & \textbf{37.97} & \textbf{88.76} & \textbf{96.83}
  & \textbf{11.15} & \textbf{40.67} & \textbf{87.51} & \textbf{96.40}
  & \textbf{13.84} & \textbf{54.63} & \textbf{84.79} & \textbf{95.13} \\
    \bottomrule
    \end{tabularx}
  }
  \vspace{-0.8em}
  \caption{Multi-layer baseline methods evaluated on our synthetic validation set. 
  Values are scaled by 100 for clearer comparison.
  Best scores are in \textbf{bold}. Second best \underline{underlined}.
  }
  \label{tab:supp_eval_syn_multi}
  \vspace{-1em}
\end{table*}

\subsection{Additional Details}

\subsubsection{Benchmark}
All images in our benchmark are released under the CC0 license. We will make the dataset publicly available; however, a subset of the ground truth annotations for the real-world images will be withheld for use on a public evaluation server.
To validate our approach, we manually annotated 30 synthetic images with known depth ground truth. Our annotations matched the ground truth in 98\% of cases, demonstrating the reliability.
In total, we annotated 1,500 images, with 300 allocated for validation and 1,200 for testing. Our annotations include 5,406 monotonic depth lines and 38,392 relative depth points across 7 distinct layers, plus 3,011 fake depth lines and 6,025 fake relative depth points.

\subsubsection{Baseline Training}
We train all NeWCRFs baseline models from scratch on our synthetic dataset for 100 epochs. For the Multi-head strategy, we set the number of output depth channels from 1 to 4 to predict multi-layer depth. For the Index Concat strategy, we concatenate the image with an index channel of size $H \times W$, where all values are set to the corresponding layer index, and set the number of input channels to 4 (RGB plus Index). For the Recurrent strategy, we concatenate the image with the previous depth output and also set the number of input channels to 4 (RGB plus Depth). During each training step, a random layer is selected as the prediction target. To provide richer supervision, we utilize snapped layered depth: if a pixel lacks ground-truth depth at layer $i$, it inherits the depth value from layer $i-1$.
For optimization, we use the Scale-Invariant Logarithmic loss~\cite{silog_loss}.

\subsubsection{Evaluation and Fine-tuning}
All methods are evaluated using a single NVIDIA RTX 3090 GPU. When assessed on the synthetic validation dataset, both the ground-truth values and predictions are clipped to the range $(0.001, 30)$. Fine-tuning for Metric3D V2 \cite{metric3dv2} is performed using the publicly available code, with training for 100,000 steps.

\subsection{Additional Results}

\subsubsection{Fine-tuning DepthAnything V2}
The proposed multi-layer depth baseline method can be applied to any existing depth backbone. To demonstrate how recent models can benefit from our dataset, we further fine-tuned DepthAnything V2 \cite{depthanythingv2} for multi-layer depth estimation using the multi-head strategy. To achieve this, we replicate the depth head four times, enabling the model to predict multi-layer depth.

\begin{figure*}[t]
    \includegraphics[width=\linewidth]{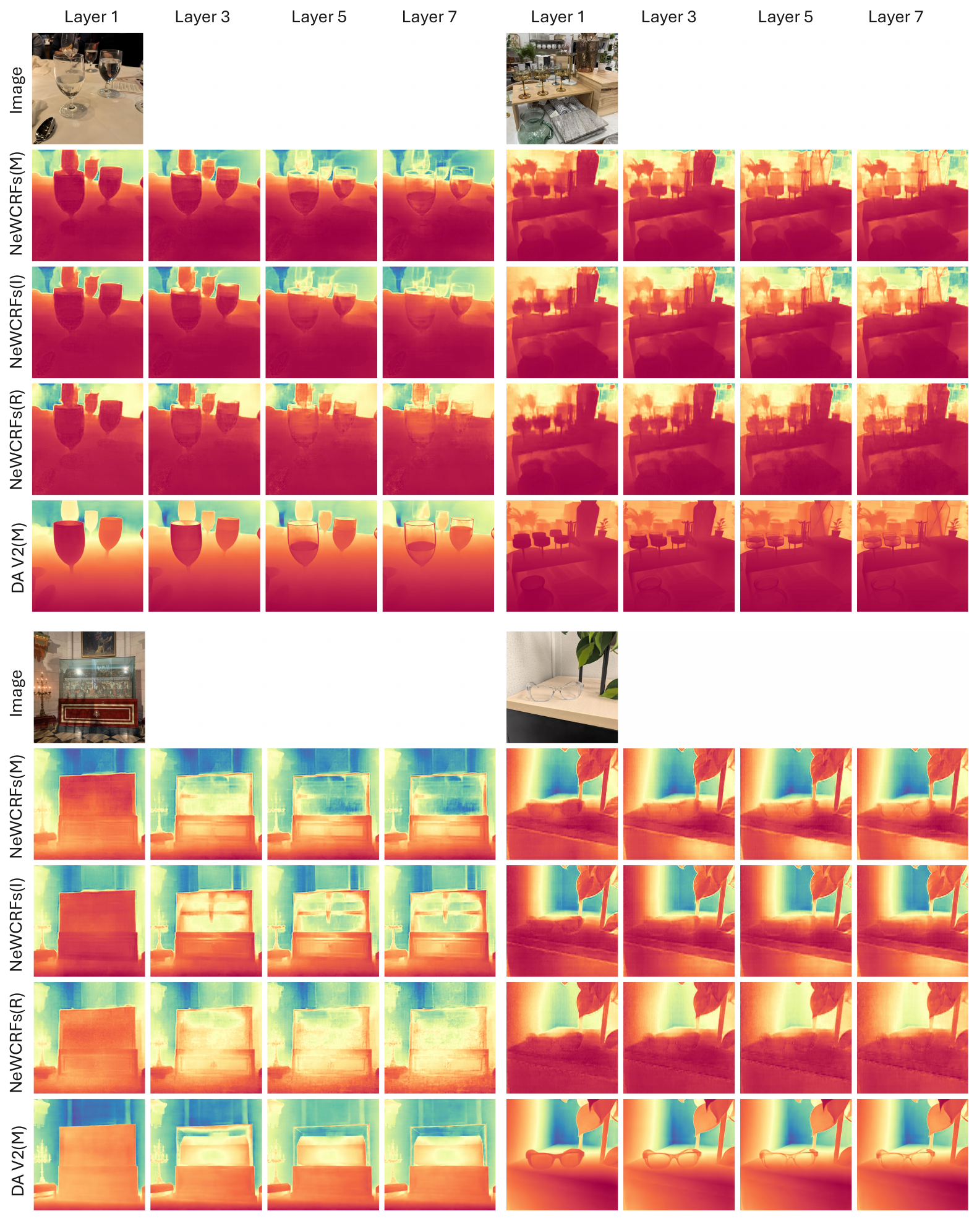}
    \caption{Additional qualitative comparison of multi-layer depth baselines on real benchmark. }
    \label{fig:layered_real1}
\end{figure*}

\begin{figure*}[t]
    \includegraphics[width=\linewidth]{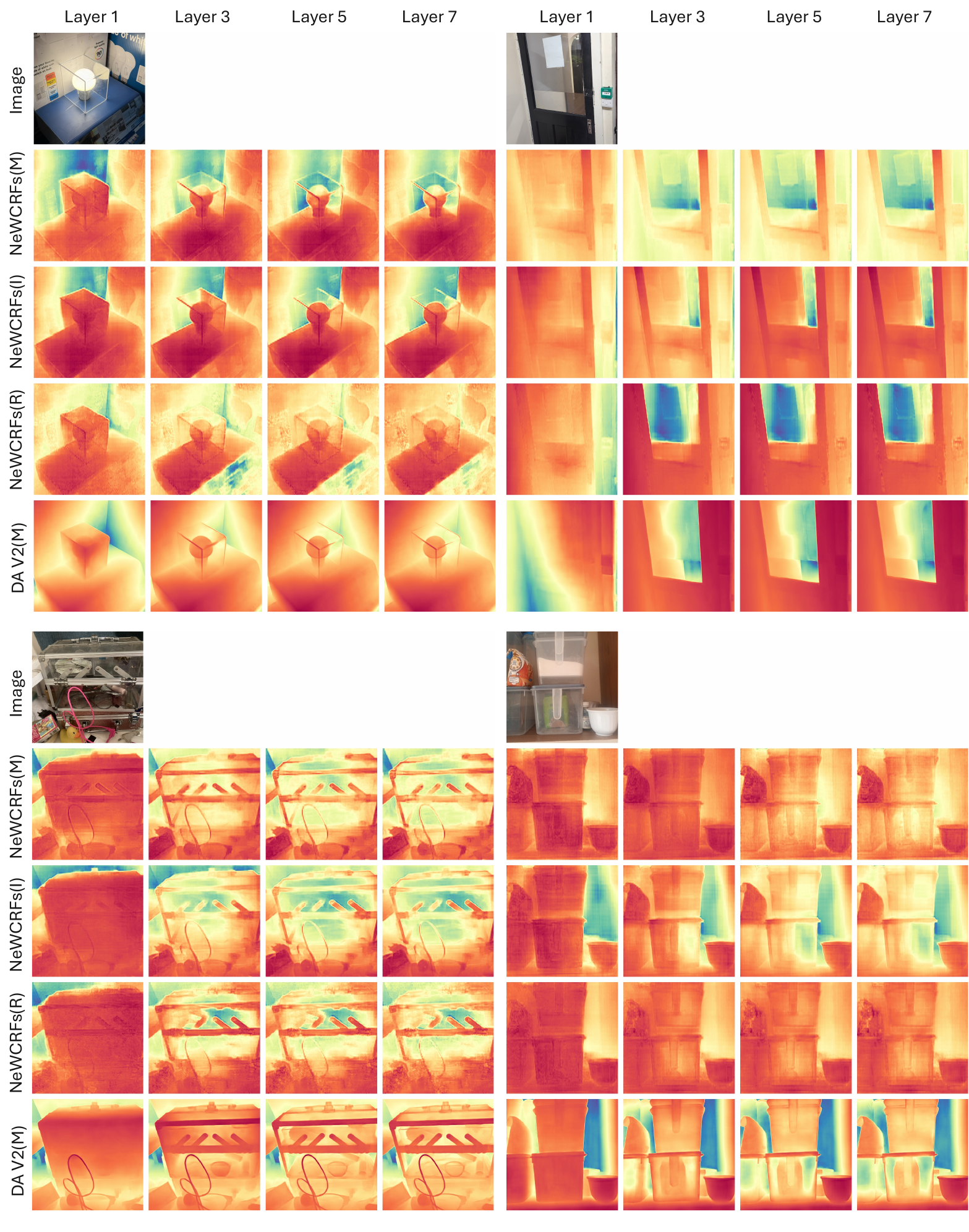}
    \caption{Additional qualitative comparison of multi-layer depth baselines on real benchmark. }
    \label{fig:layered_real2}
\end{figure*}

\begin{figure*}[t]
\centering
    \includegraphics[width=0.85\linewidth]{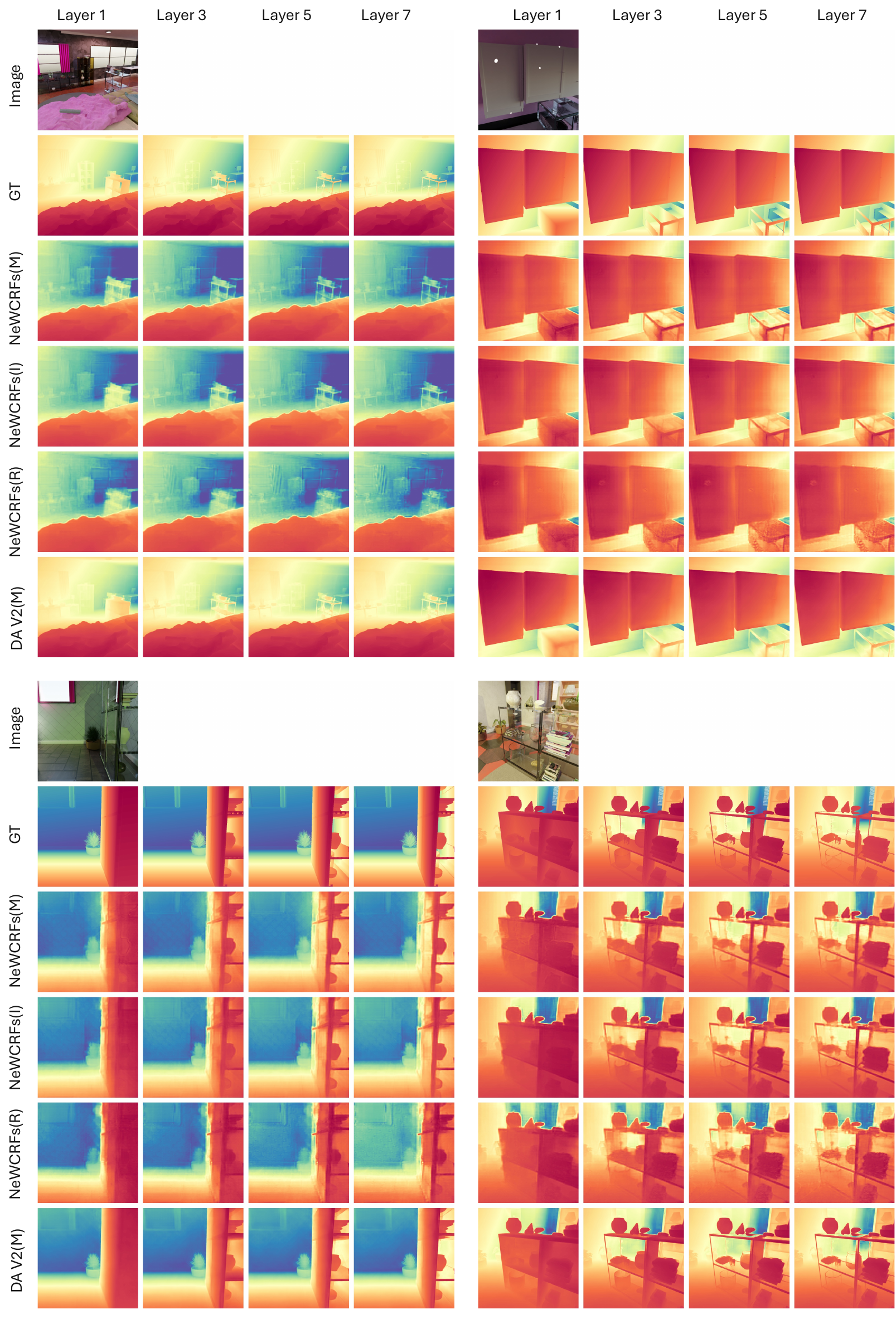}
    \caption{Additional qualitative comparison of multi-layer depth baselines on synthetic validation set. }
    \label{fig:layered_syn}
\end{figure*}

Results evaluated on the real world benchmark are shown in \cref{tab:supp_eval_real_baseline}. Results evaluated on the synthetic validation set are shown in \cref{tab:supp_eval_syn_multi}. Qualitative comparison on real benchmark can be found in \cref{fig:layered_real1} and \cref{fig:layered_real2}. Qualitative comparison on synthetic validation set can be found in \cref{fig:layered_syn}.
These results show that DepthAnything V2 not only aligns more closely with the training domain but also generalizes exceptionally well after fine-tuning solely on our synthetic data. This underscores the power of our synthetic data generator for multi-layer depth estimation and transparent-object understanding. Nevertheless, some regions around object boundaries and in deeper layers still exhibit artifacts, leaving rooms for improved model designs in future work.

\subsubsection{Additional Qualitative Results}
\begin{figure*}[t]
    \includegraphics[width=\linewidth]{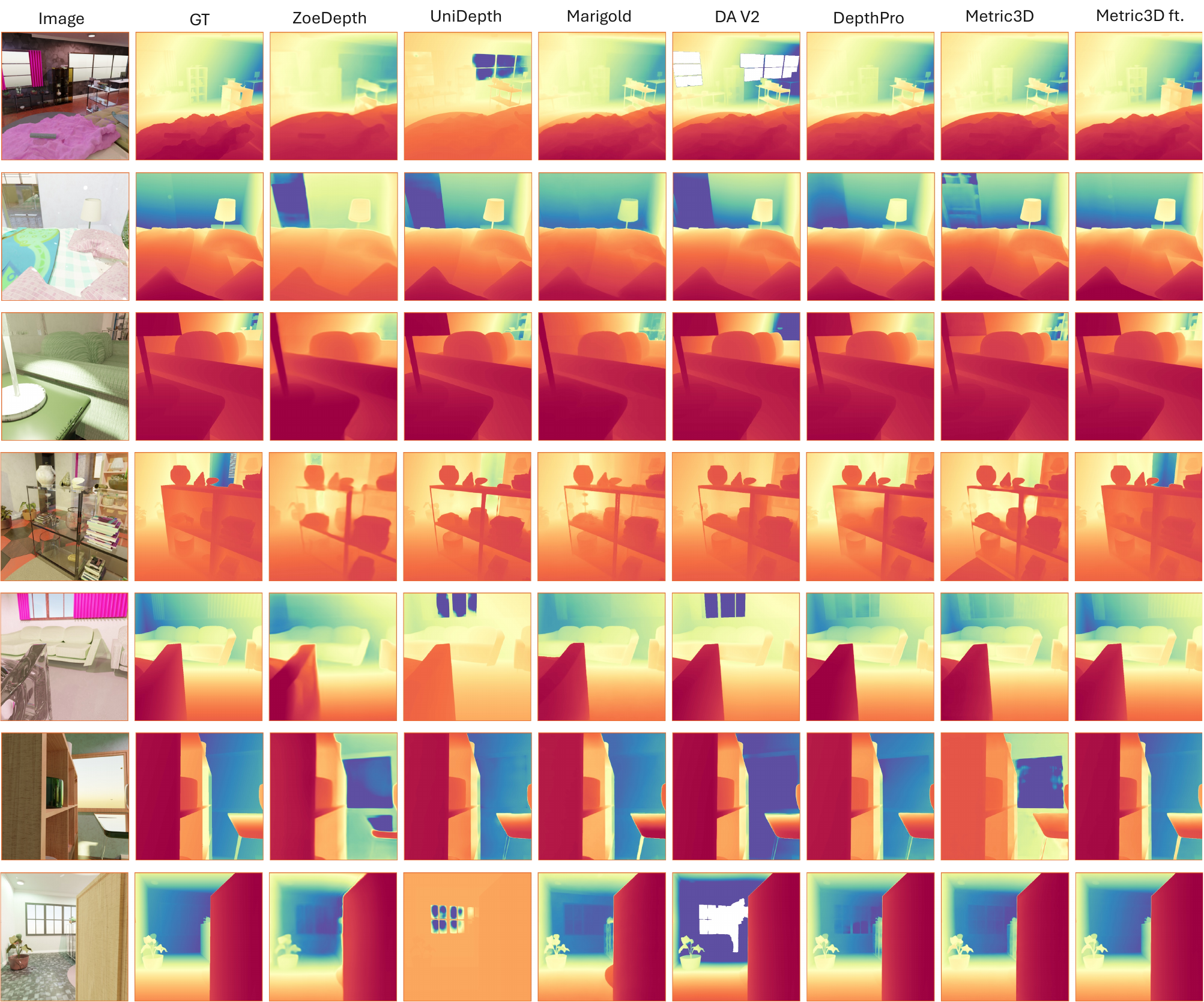}
    \caption{Additional qualitative comparison of state-of-the-art single layer depth methods on our synthetic validation set. }
    \label{fig:syn_firstlayer}
\end{figure*}

To help readers better visually assess the effectiveness of our datasets,
we provide additional qualitative results on real world benchmark and synthetic validation set, which can be found in \cref{fig:realworld_firstlayer} and \cref{fig:syn_firstlayer}, respectively.

\begin{figure*}[t]
    \includegraphics[width=\linewidth]{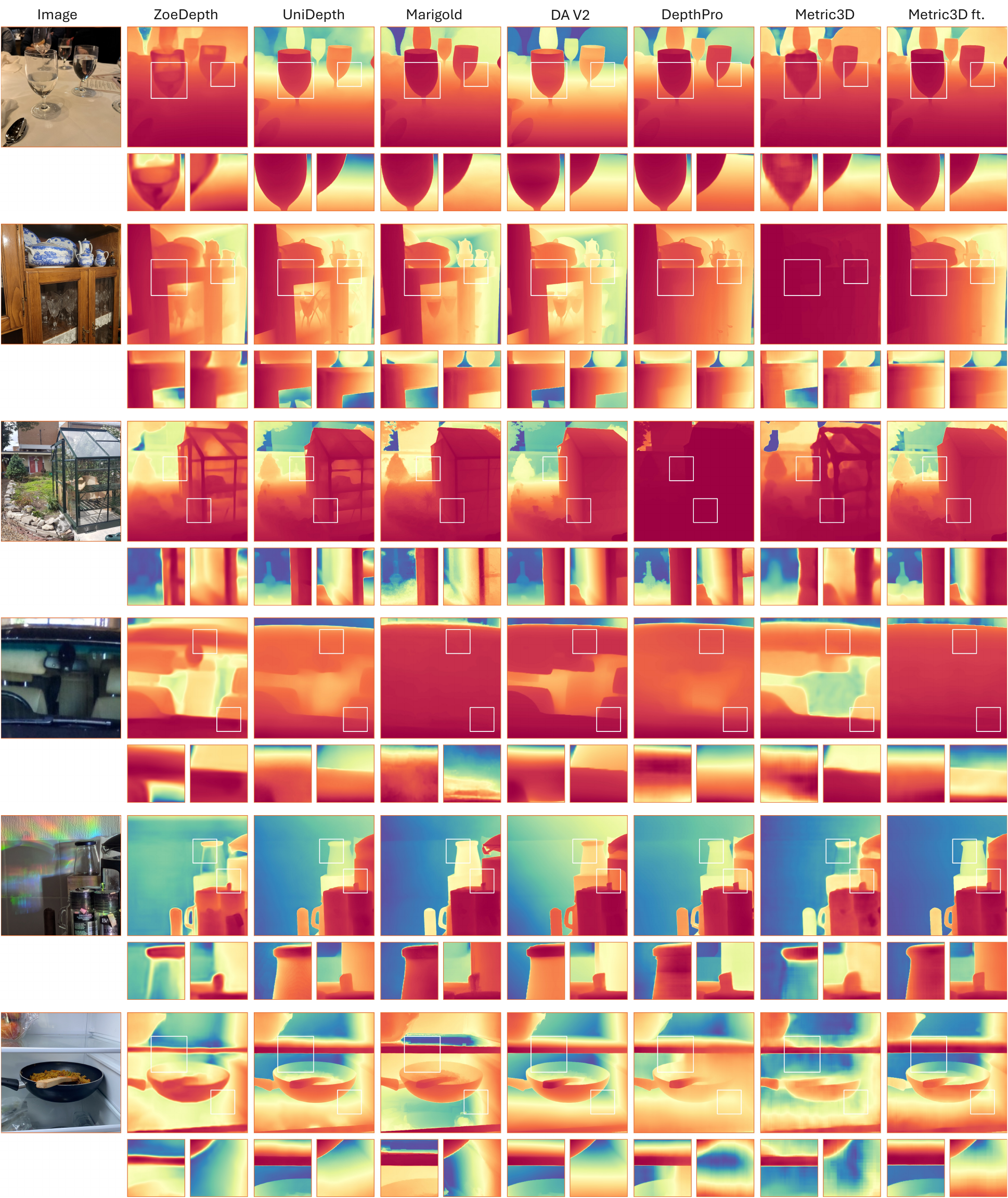}
    \caption{Additional qualitative comparison of state-of-the-art single layer depth methods on our benchmark.}
    \label{fig:realworld_firstlayer}
\end{figure*}

\clearpage
\bibliographystyle{ieeenat_fullname}
\bibliography{main}
\end{document}